\documentclass{article}

\PassOptionsToPackage{square,sort,comma,numbers}{natbib}

\usepackage[preprint]{neurips_2021}

\usepackage[utf8]{inputenc} 
\usepackage[T1]{fontenc}    
\usepackage{hyperref}       
\usepackage{url}            
\usepackage{booktabs}       
\usepackage{amsfonts}       
\usepackage{nicefrac}       
\usepackage{microtype}      
\usepackage{xcolor}         
\usepackage{wrapfig}
\usepackage{subfig}
\usepackage{amsmath}
\usepackage{graphicx}
\usepackage[export]{adjustbox}
\usepackage{algorithm}
\usepackage[noend]{algpseudocode}
\usepackage{caption}
\usepackage[normalem]{ulem}

\title{A coarse space acceleration of deep-DDM}

\author{%
  Valentin~Mercier \\
  \thanks{IRIT Computer Science Research Institute of Toulouse,\texttt{valentin.mercier@etu.enseeiht.fr}}
  \And
  Serge Gratton \\
  \thanks{IRIT Computer Science Research Institute of Toulouse,\texttt{serge.gratton@toulouse-inp.f}}
  \AND
  Pierre Boudier \\
  \thanks{Nvidia}
}

\begin{document}
\maketitle

\begin{abstract}
The use of deep learning methods for solving PDEs is a field in full expansion. In particular, Physical Informed Neural Networks, that implement a sampling of the physical domain and use a loss function that penalizes the violation of the partial differential equation, have shown their great potential. Yet, to address large scale problems encountered in real applications and compete with existing numerical methods for PDEs, it is important to design parallel algorithms with good scalability properties. In the vein of traditional domain decomposition methods (DDM), we consider the recently proposed deep-ddm approach. We present an extension of this method that relies on the use of a coarse space correction, similarly to what is done in traditional DDM solvers. Our investigations shows that the coarse correction is able to alleviate the deterioration of the convergence of the solver when the number of subdomains is increased thanks to an instantaneous information
exchange between subdomains at each iteration. Experimental results demonstrate that our approach induces a remarkable acceleration of the original deep-ddm method, at a reduced additional computational cost.
\end{abstract}

\section{Introduction}
A large part of the known physical phenomena are modeled by partial differential equations (PDE). Given that most of them do not have analytical solutions, their resolution by numerical methods is a major challenge in applied mathematics. With the reemergence of deep learning techniques for computer vision problems, the use of neural networks has expanded into many areas. Their ability to extract information from large databases and their approximation capabilities have made them formidable candidates for solving many problems. It is for these reasons that hybridization between deep learning models and numerical simulation has multiplied \cite{von_rueden_combining_2020}. Similarly, techniques for integrating physical knowledge into learning processes have become more widespread \cite{von_rueden_informed_2020}. 

Successful examples of this new approach include the so-called Physics informed neural networks (PINN) \cite{raissi_physics-informed_2019}. They allow the resolution of PDEs on a given domain using only the equations of the physical problem. They are mesh-free and use only sampled points within and at the boundaries of the domain. Their ability to approximate a given physics is based on an integration of the problem equations into the loss function. 

PINNs are empirically effective and have been adapted to various physical problems (see \cite{rao_physics-informed_2020} or \cite{kadeethum_physics-informed_2020}). A theoretical justification for their convergence is given in \cite{shin_convergence_2020}. As detailed in the latter paper, the convergence (for a defined model space) of PINNs depends on the amount of sampled points taken inside and at the boundaries of the domain. However, over a large domain and for complex physical situations, the cost of training these networks can easily become problematic in view of the amount of sampling points needed to reach satisfactory accuracy on the solution.  This difficulty has already been dealt with and overcome in numerical methods for solving PDEs. A successful approach, that originates from the work of Hermann Schwarz from the nineteenth century, relies on a 'divide and conquer' strategy, that consists in splitting our problem into sub-problems of adequate size (\cite{toselli_domain_2005}). By analogy, the idea of using small networks on sub-domains seems to be a ready-made solution to address the computational complexity of training PINNs. 

From this observation, it seems natural to incorporate  domain decomposition techniques into the PINN approach.  This solution will have a double advantage: it will reduce the cost of training each PINN on its sub-domain and it will naturally lead to methods amenable to  distributed computing. This work has been initiated by Li et al.~\cite{li_deep_2020} where the link between the classical additive Schwartz method and the PINN concept is made.

Before further discussing~\cite{li_deep_2020}, it is worth briefly outlining alternative approaches combining DL and DD. For example, \cite{li_d3m_2020} presents the deep-ddm approach, in which  Ritz-Galerkin networks \cite{e_deep_2017} are used, instead of PINNs. These networks rely on a variational form of the equations in their training phase. Other methods listed in \cite{heinlein_combining_2021} use ML in preparatory phases of DD algorithms to improve their convergence properties. Another approach to the parallelization of PINNs using non-overlapping methods is presented in \cite{jagtap_conservative_2020} and \cite{karniadakis_extended_2020}. Although these methods are parallelizable (see \cite{shukla_parallel_2021}) they are quite different from Schwarz methods which only exchange their interface conditions: here the interface conditions are coupled at each time. 

In~\cite{li_deep_2020}, contrary to standard DDM approaches for, say, elliptic PDEs, a neural network approximation is use in place of  the (direct or iterative) sub-domains solution. Empirically, the deep-ddm method follows the theoretical convergence properties of the classical methods while producing a continuous result over the entire domain. The main limitation of the method are as expected those of classical Schwarz methods. The transmission of the information along the iteration only occurs between neighboring subdomains. As a result, the speed of convergence of the method (i.e. the number of iterations to reach a given tolerance) deteriorate as the number of subdomains increases: the methods does not scale well  when processor count is increased \cite{bercovier_development_2009}. 

In standard DDM, this pitfall is avoided with the addition of a so-called coarse grid correction allowing for a fast transfer of information accross the whole domain. Although classical in the state of the art of domain decomposition methods \cite{dolean_introduction_2015}, to the best of our knowledge, this technique has not  been developed with a DL vision yet.

In this article we will try to answer the following questions: 
How to introduce a mechanism similar to the coarse grid into a deep-ddm method? The fact that the accuracy obtained with neural networks on the coarse problem and on the subdomains may not compare favorably with that obtained in DDM, will the introduction of a coarse grid still be computationally interesting ? 

Throughout this article we will be interested in elliptical problems of the following form: 

Let $\Omega \subset \mathbf{R}^d$ a domain and $\partial \Omega$ it boundary :
\begin{equation}
\label{general_problem}
    \left\{
    \begin{aligned}
    &\mathcal{L}(u)= f  \text{ in } \Omega  \\
    &\mathcal{B}(u) = g  \text{ on }  \partial \Omega     
    \end{aligned}
  \right.
\end{equation}
where $\mathcal{L}$ and $\mathcal{B}$ are two differential operators  and $f$ and $g$ are two given functions. 

With this general notation, we can for example write the Poisson problem with Dirichlet boundary conditions by taking the Laplace operator for $\mathcal{L}$ and the identity map for $\mathcal{B}$.

\section{Domain Decomposition Méthods and PINN}
\subsection{Basics of domain decomposition}
Domain decomposition methods originate from the work of Hermann Schwarz \cite{schwarz_ueber_1870} on the solution of the Poisson equation. Before this work, the existence of a solution was only proven for simple geometries. Hermann Schwarz had the idea to consider a domain of complex geometry as a union of simple geometries as described in Figure~\ref{fig:dd}.From there he sets up an iterative algorithm: local solutions of the Poisson equation are performed on subdomains taking for boundary condition the value of the solution on the neighboring subdomains. In the case of $2$ subdomains, Schwarz's methods writes:

\begin{figure}[hbtp]
 \centering
  \includegraphics[width=4cm]{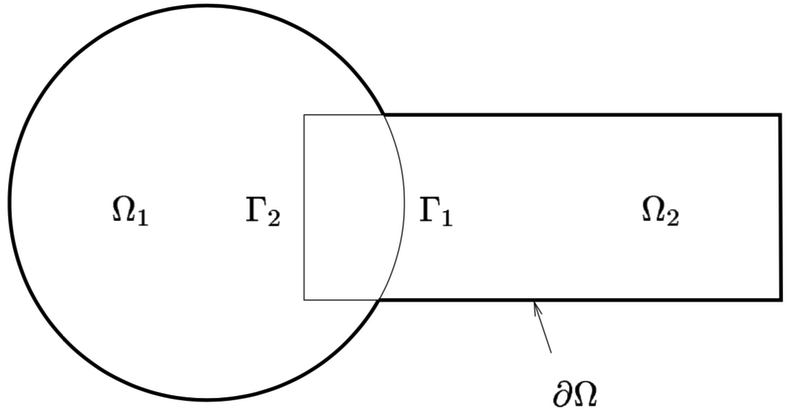}
  \caption{The diagram used by Schwarz in 1870.}\label{fig:dd}
\end{figure}

\begin{equation}
\label{Schwarz_alternating_method}
\left\{
\begin{aligned}
&\Delta u_1^{k+1} = f \text{ in } \Omega_1\\
&u_1^{k+1} = g \text{ on } \Gamma \cap \partial \Omega_1\\
&u_1^{k+1} = u_2^{k} \text{ on } \Gamma_1\\
\end{aligned}\right.
\quad \mbox{ and } \quad
\left\{\begin{aligned}
&\Delta u_2^{k+1} = f \text{ in } \Omega_2\\
&u_2^{k+1} = g \text{ on } \Gamma \cap \partial \Omega_2\\
&u_2^{k+1} = u_1^{k+1} \text{ on } \Gamma_2.\\
\end{aligned}\right. 
\end{equation}

After showing the convergence of this algorithm he obtains, in the limit, a solution of the Poisson equation on the composite geometry. 
With the emergence of numerical simulation, these methods have regained interest because they can me modified into a parallel algorithm. Indeed, with few modifications, we can obtain an algorithm whose solutions on the subdomains can be perfomed concurrently as follows: 
\begin{equation}
\label{Additive_Schwarz_method}
\left\{
\begin{aligned}
&\Delta u_1^{k+1} = f \text{ in } \Omega_1\\
&u_1^{k+1} = g \text{ on } \Gamma \cap \partial \Omega_1\\
&u_1^{k+1} = u_2^{k} \text{ on } \Gamma_1\\
\end{aligned}\right.
\quad \mbox{ and } \quad
\left\{\begin{aligned}
&\Delta u_2^{k+1} = f \text{ in } \Omega_2\\
&u_2^{k+1} = g \text{ on } \Gamma \cap \partial \Omega_2\\
&u_2^{k+1} = u_1^{k} \text{ on } \Gamma_2.
\end{aligned}\right. 
\end{equation}
Note the method we described so far is mainly theoretical, as it involves the solutions of continuous problems. To solve numerically the problem it is necessary to apply a discretization scheme on each subdomain, using for example a method such as the the finite element method. 
The central idea of the article \cite{li_deep_2020} is to replace these numerical solvers by neural networks and more precisely by PINN.
\subsection{About PINN}
Physics Informed Neural networks constitute  a class of neural networks dedicated to the resolution of PDEs~\cite{raissi_physics-informed_2019}. Consider problem (\ref{general_problem}), and let us denote by $h$ a fully connected neural network of $L$ hidden layers of $I$ neurons with differentiable activation functions $o$, and by $\theta$ the parameters of our network. We want to train the network so that 
$$ h(x, \theta) \sim u \quad \forall x \in \Omega , $$
where $u$ denote the solution of problem~(\ref{general_problem}).
We introduce a sample of points inside $\Omega$ ($X_f=\{x^i_f\}_{i=1}^{N_f}$) and at its boundary ($X_g=\{x^i_g\}_{i=1}^{N_g}$) of the domain. Taking advantage of the differentiable nature of the network the following problem is introduced: 
\begin{equation}
\begin{aligned}
    &\theta^* = \underset{\theta}{\mbox{argmin}}\,  \mathcal{M}(\theta),  \mbox{ where }
 \end{aligned}
 \end{equation}
 \begin{equation}
 \left \{
 \begin{aligned}
    &\mathcal{M}(\theta) = \mathcal{M}_\Omega(\theta) + \mathcal{M}_{\partial \Omega}(\theta), \\
    &\mathcal{M}_\Omega(\theta) = \frac{1}{N_f}\sum_{i=1}^{i=N_f}|\mathcal{L}({h(x_f^i)})-f(x_f^i)|^2,\\
    &\mathcal{M}_{\partial \Omega}(\theta) = \frac{1}{N_g}\sum_{i=1}^{i=N_g}|\mathcal{B}({h(x_g^i)})-f(x_g^i)|^2.\\
 \end{aligned}
 \right .
  \end{equation}

In this formulation, the available physical knowledge is encoded as a penalty term in the loss function of the neural network. 
A study of PINN convergence has been conducted in \cite{shin_convergence_2020}  where it is shown that the accuracy of the method (and thus of the physical solution) essentially relies $3$ points: on the number of sampling points, on the convergence reached in the training phase, and on the error of approximation of the true function by the considered  neural network architecture. 

Let us consider the use of PINN solvers in a DD approach.
\subsection{Deep-ddm}
The deep-ddm~\cite{li_deep_2020} algorithm is based on the domain decomposition algorithm described in (\ref{Additive_Schwarz_method}). 
Let S be the number of subdomains, let $*_s$ be the elements associated with the subdomain $s (1<s<S)$ and $*_r$ the elements associated with the corresponding neighbour of $s$. 

We consider the local problems:
\begin{equation}
\label{local_problem}
    \left\{
    \begin{aligned}
    &\mathcal{L}(u_s)= f  \text{ in } \Omega_s  \\
    &\mathcal{B}(u_s) = g  \text{ on }  \partial \Omega_s \backslash \Gamma_s \\
    &\mathcal{D}(u_s) = \mathcal{D}(u_r)  \text{ on }  \Gamma_s
    \end{aligned}
  \right.
\end{equation}
with $\mathcal{D}$ the operator of an artificial interface transmission condition and $\Gamma_s$ the interfaces between a sub-domain and its neighbours.

The PINNs presented in the previous section are then slightly modified by an additional loss term related to the interfaces between the domains. By denoting $X_\Gamma = \{x^i_\Gamma\}_{i=1}^{N_\Gamma}$ the sampled points on the interfaces and $h_r$ the corresponding neighbouring network we add to the total loss the term 

\begin{equation}
\begin{aligned}
    \mathcal{M}_\Gamma(\theta) &= \frac{1}{N_\Gamma}\sum_{i=1}^{i=N_\Gamma}|\mathcal{D}({h_s(x_\Gamma^i)})- W^i_s|^2\\
    W^i_s &= \mathcal{D}({h_r(x_\Gamma^i)}).
\end{aligned}
\end{equation}

Let $X_s= {X_{fs},X_{gs},X_{\Gamma s}}$ the sampled points on the sub-domain $s$, the deep-ddm pseudocode writes: 

\begin{algorithm}[!h]
\caption{DeepDDM for the s-th subproblem}\label{alg:deep-ddm}
\begin{algorithmic}[1]
\State Sampling of the $X_s$ points
\State Initialization of the network parameters $\theta_s^0$
\State Initialization of information at interfaces $W_s^0$
\While{Non convergence and iteration limits not reached}
\State Local network training
\State Update of values at interfaces 
\State Network convergence test 
\State Interface convergence test
\EndWhile
\end{algorithmic}
\end{algorithm}

There are two stopping criteria in the deep-ddm algorithm, the first one in line 7 concerns the convergence of networks. If the relative norm between the prediction of $h_s(\theta_s^k)$ and $h_s(\theta_s^{k+1})$ on the interior points $X_{fs}$ is less than some tolerance threshold, then the network has converged. The second (line 8) tests the convergence of the interfaces, let $X_\Gamma$ be the points of an interface $h_r$ the neighbouring network covering that interface, if the relative norm $h_r(\theta_r^k)(X_\Gamma)$ and $h_r(\theta_r^{k+1})(X_\Gamma)$ is less than a tolerance threshold then the interfaces have converged. If all interfaces or all networks have converged then the algorithm stops. 
For further details on the training conditions of the network, we refer to~\cite{li_deep_2020}. 

\subsection{Schwarz additive limits and scalability }

As shown in~\cite{li_deep_2020}, the deep-ddm method has similar convergence properties as a classical additive Schwarz method with finite elements, and suffers from the same drawback. In particular, the performance of the Schwarz algorithm deteriorates as  the number of subdomains growth, which is refereed to as a lack of {\em scalability} in the literature~\cite{dolean_introduction_2015}. 
More precisely two types of scalability are used in the evaluation of performance of parallel algorithms : weak scalability and strong scalability~\cite{dolean_introduction_2015}: 
\begin{itemize}
    \item \emph{Strong Scalability :} Strong scalability is defined as how the solution time varies with the number of cores for a fixed total problem size. Ideally, the elapsed time is inversely proportional to the number of processing units (CPU's,cores,GPU's). 
    \item \emph{Weak Scalability :} Weak scalability is defined as how the solution time varies with the number of cores for a fixed problem size per core. Ideally, the elapsed time is constant for a fixed ratio between the size of the problem and the number of processing units (CPU's,cores,GPU's). 
\end{itemize}

Strong scalability is of particular interest to us because it allows us to scale the computing power needed to solve a given problem within a prescribed computational time. Let us investigate on the strong scalability of the deep-ddm method.

Note that unlike \cite{shukla_parallel_2021}  we measure our scalability in terms of iterations needed to achieve a convergence threshold (with relative error) and not in terms of solution time or data points/sec related to a given parallel implementation. Such a more refined analysis of a parallel implementation will be the scope of a future work.  For the time being, and to provide a reference that will enable us to assess the improvements obtain with our coming algorithmic variant, we  now propose a brief experiment on the strong scalability of our implementation.

In our simulation of  a parallel method, since the number of cores is represented by the number of subdomains used, we increase this number for a fixed given continuous problem. In  our experiment, a given  number of points is evenly distributed across the  subdomains. The neural architecture is taken identical in each subdomain, and does not depend on the number of subdomains. A detailed description of the experience is given in the supplementary material. 


\begin{wrapfigure}{R}{0.5\linewidth}
    \centering
    \includegraphics[width=\linewidth]{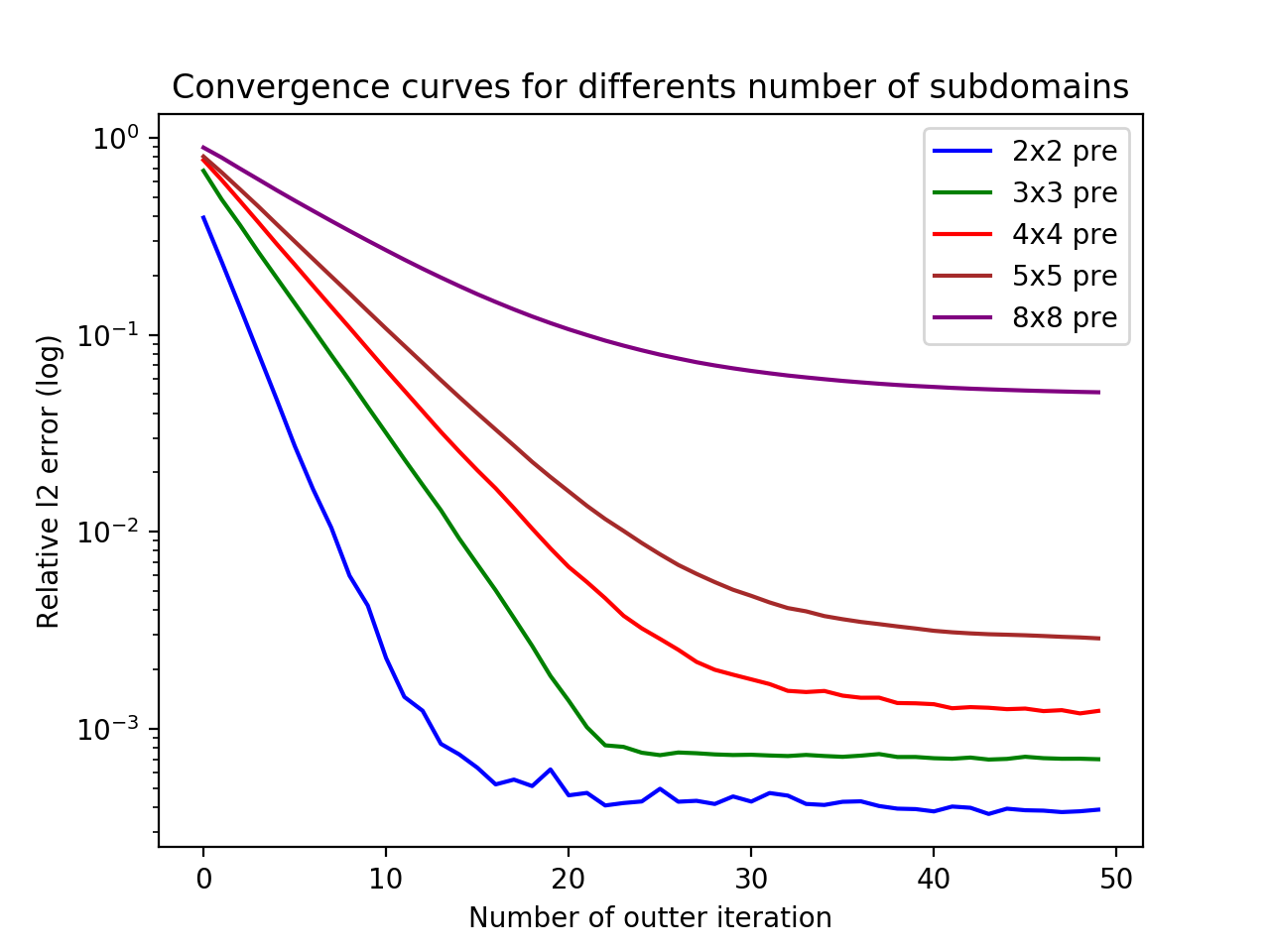}
    \caption{Strong scalability test on deep-ddm}
    \label{nonscal}
\end{wrapfigure}
The graph in figure \ref{nonscal} is obtained by comparing the Deep-ddm solution with the analytical solution of the problem. We observe a scalability issue analogous to that of classical Schwarz methods. This can be explained by the way information is exchanged in the course of iterations. Each local resolution only communicates with its neighbors, using values at the interfaces. Thus, information from a subdomain $S1$ will take as many iterations to reach a subdomain $S_n$ as there are subdomains between them.  
In the state of the art of DDM, in order to improve the scalability, so-called two-level methods are employed; they use a coarse space to transmit information instantaneously throughout the computational domain \cite{dolean_introduction_2015}. In a similar way we propose now to implement this concept in the deep-ddm algorithm.



\section{Coarse problem and Deep DDM}
Two-level DD methods involve a decomposition into local sub-problems and a global coarse representation of the problem. In our framework, each  (sub-)domain is associated with sampling points laying in its interior and on its boundaries. We will accordingly associate with our coarse space a (coarse) sampling of our complete domain. As a first attempt, we associate to these points a fully-connected network (with hyperbolic tangent as activation function) of the same dimension as those associated to local resolutions. We shall now describe  the connection we implement between local sub-problems and the (global) coarse problem.  
\subsection{The coarse problem}
\label{global:pb}
Let the points of the coarse problem be $X^{coarse} = \{X^{coarse}_f,X^{coarse}_g\}$ with $x_f^{i,coarse} \in X^{coarse}_f$ the interior points and $x_g^{i,coarse} \in X^{coarse}_g$ the boundary points. We denote by $h_{coarse}$ the coarse neural network and by $\theta_{coarse}$ its parameters. To  write algorithms that act on global functions, we now introduce extension and unit partition operators. 

\emph{Extension operators and a partition of unity~\cite{dolean_introduction_2015} :} Let the extension operator $E_i$ be such that $E_i(w_i):\Omega \rightarrow \mathbb{R}{}$ is the extension of a function $w_i : \Omega_i \rightarrow \mathbb{R}{}^n$, by zero outside $\Omega_i$. We also define the partition of unity functions : $\chi_i:\Omega \rightarrow \mathbb{R}{}^n,\chi_i\geq 0$, and $\chi_i(x)=0$ for $x \in \partial \Omega_i \backslash \partial \Omega$ and such that :
$$ w = \sum_{s=1}^S E_i(\chi_i w_{|\Omega_i})$$
for any function $w : \Omega \rightarrow \mathbb{R}{}^n$.

We can define the optimisation problem associated with our coarse network as   
\begin{equation*}
\begin{aligned}
    \theta_{c}^* &= \underset{\theta_{c}}{\mbox{argmin}} \mathcal{M}(\theta_c) = \mathcal{M}_\Omega(\theta_c) + \mathcal{M}_{\partial \Omega}(\theta_c) +  \lambda_f  \cdot \mathcal{M}_{fine}(\theta_c), \mbox {where }\\
\end{aligned}
\end{equation*}
\begin{equation}
\left \{
\begin{aligned}
    \mathcal{M}_\Omega(\theta_c) &= \frac{1}{N_{f,coarse}}\sum_{i=1}^{i=N_{f,coarse}}|\mathcal{L}({h_{coarse}(x_f^{i,coarse})})-f(x_f^{i,coarse})|^2\\
    \mathcal{M}_{\partial \Omega}(\theta_c) &= \frac{1}{N_{g,coarse}}\sum_{i=1}^{i=N_{g,coarse}}|\mathcal{B}({h_{coarse}(x_g^{i,coarse})})-f(x_g^{i,coarse})|^2\\
    \mathcal{M}_{fine}(\theta_c) &= \frac{1}{N_{f,coarse}}\sum_{i=1}^{i=N_{f,coarse}}|h_{coarse}(x_f^{i,coarse}) - \sum_{s=1}^S E_i(\chi_i h_i(x_f^{i,coarse}))|^2.\\
\end{aligned}
\right .
\end{equation}

We use the loss function $\mathcal{M}_{fine}(\theta)$ to transfer the information from the fine model to the coarse model. More precisely, the coarse model is constrained "to stay close" to the values of the fine resolution at the interior points of the coarse problem.  This is implemented with a penalty parameter $\lambda_f$ in front of this loss function.  This is a way to monitor the impact of the sub-domain solution (which may be approximate at early stages of the iterations)  to the coarse problem.  

The training of the coarse network is similar to that of the local networks described in \cite{li_deep_2020}. A new parameter $tol_m^{coarse}$ will be defined for this training. As a reminder, the training is stopped when the maximum number of iterations is reached or if the loss function is sufficiently stabilised, i.e. if : 
$$ \frac{\mathcal{M}(\theta_c^i) -\mathcal{M}(\theta_c^{i-\eta})}{\mathcal{M}(\theta_c^i)} \leq tol_m^{coarse} \text{~~ where $\eta$ is a definite integer.} $$
\subsection{The fine problem}
We keep for our fine problem the same elements (networks, sampled points) as those used in the deep-ddm algorithm. In the deep-ddm algorithm, we have forced the local network $h_s$ to take (through the transfer operator of our physical problem) 
the value obtained by its neighbouring network $h_r$ that is defined by  $W_s^i = \mathcal{D}({h_r(x_{\Gamma}^i)})$ . With the introduction of the coarse  information we now define the value  at the interface as a combination of these two values by
$$ W_s^i = \lambda_c \cdot \mathcal{D}({h_{coarse}(x_\Gamma^i)})+ (1-\lambda_c) \cdot \mathcal{D}({h_r(x_\Gamma^i)}), $$
with $\lambda_c$, $0\leq \lambda_c \leq 1$, weighting the influence of the coarse network on the information at the interfaces.
\subsection{Algorithm}
The sub-problems and coarse problems being described, we are now in position to describe our two-level method in Algorithm 2.
\begin{center}
\setlength{\textfloatsep}{0pt}
\setlength{\floatsep}{0pt}
\setlength{\intextsep}{0pt}
\begin{algorithm}[htp]
\caption{Two-level DeepDDM}\label{alg:two-level-deep-ddm}
\begin{algorithmic}[1]
\State Sampling of the $X_s$ and $X^{coarse}$ points
\State Initialization of the network parameters $\theta_s^0$ and $\theta_c$
\State Initialization of information at interfaces $W_s^0$
\State Initialization of parameters $\lambda_f$ and $\lambda_c$
\While{Non convergence and iteration limits not reached}
\State Local network training
\State Compute $\sum_{s=1}^S E_i(\chi_i h_i(x_f^{i,coarse})$ for each coarse points in $X_f^{coarse}$
\State Coarse network training
\State Update of values at interfaces 
\State Network convergence test 
\State Interface convergence test
\State Update $\lambda_f$ and $\lambda_c$
\EndWhile
\end{algorithmic}
\end{algorithm}
\end{center}
\section{Numerical results}
\subsection{Training Procedure}
The neural  networks we consider are fully connected networks with $h$ hidden layers of $w$ neurons each, and use {\tt tanh} activation functions. Coarse networks have the same number of hidden layers and neurons as sub-domain networks; this choice is of course questionable and will be investigated in a future work. Let us just note that it keeps the computational cost of the coarse problem solution under control. The parameters of the (local and coarse) networks are trained at each outer iteration with an Adam \cite{kingma_adam_2017} optimizer until convergence of the loss functions is reached, or until maximum iteration limits is reached. 
In a first step we will experiment on a Poisson equation with Dirichlet boundary conditions. We will consider problems with known analytical solution and which enables the use the relative error $\mathcal{L}^2$ for performance analysis.

Details of each problem and all hyperparameters are given in the supplementary material . 
\subsection{Coarse network influence}
We test the influence of the coarse network on the convergence of our algorithm. To do this, we vary the $tol_m^{coarse}$ parameter, which represents the stopping criterion of our coarse network training (\ref{global:pb}). By lowering this parameter, we increase the number of epochs performed on the coarse grid and thus the quality of our coarse network is improved, making however the overall method more expensive. 
\begin{figure}[H]
    \centering
    \includegraphics[width=\linewidth]{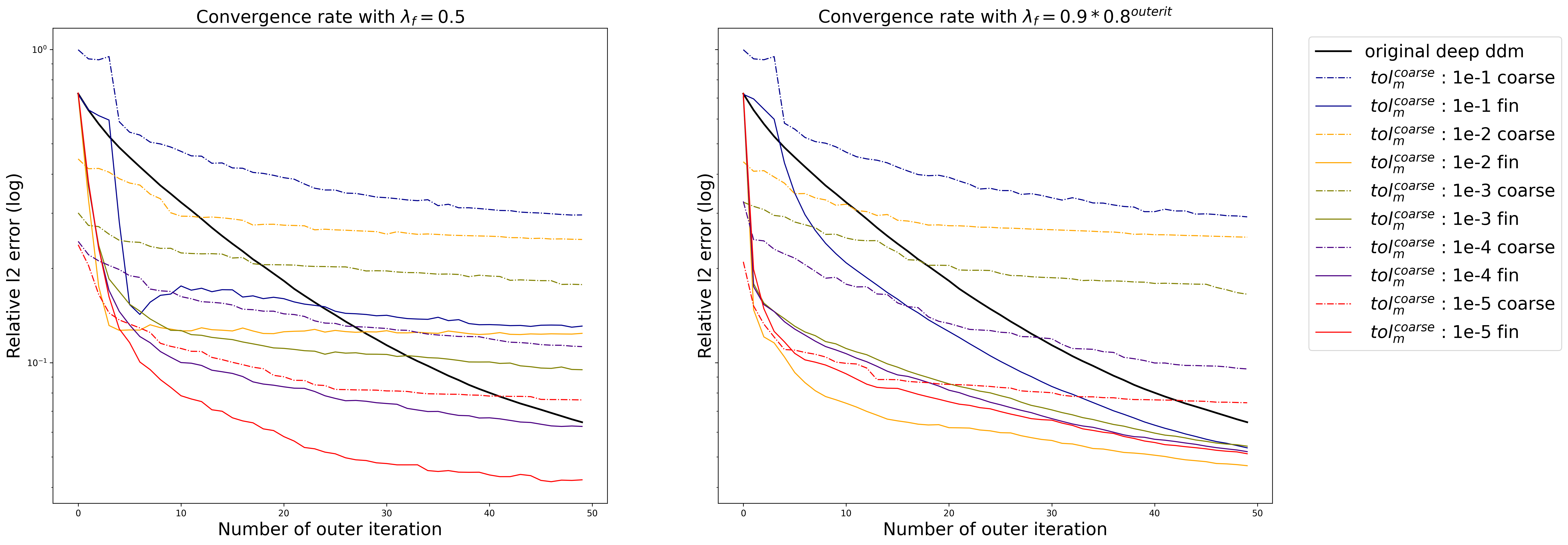}
    \caption{Test on the influence of the coarse network by varying its training criterion. On left the participation of the coarse network to the composition of the values at the interfaces is constant $\lambda_c = 0.5$, on the right it varies $\lambda_c = 0.9\cdot 0.8^{outerit}$}
    \label{fig:coarseinfluence}
\end{figure}
On the first graph, we run our algorithm with a constant parameter $\lambda_f = 0.5$; we observe in the first iterations an acceleration of the convergence, in correlation with the quality of the coarse network. After few iterations this phenomenon loses momentum and the convergence slows down; we suspect that at this point, the sub-domain networks are penalized by the contribution of the coarse network. This motivates the introduction of a dynamic adjustment of $\lambda_f$ along the iterations. We  propose a strategy where a parameter $\lambda_f$ initialized to $0.9$ with a decrease rate of $0.8$ at each iteration, this choice allows us to keep the convergence acceleration properties of the first iterations while mitigating the negative influence of the coarse network, when it occurs. This evolution scheme for $\lambda_c$ is used for all the coming numerical experiments. 

\subsection{Scalability test}
\subsubsection*{Weak scalability}
By using a fixed number of sampling points per subdomain and thus per computational unit, we investigate on the weak scalability of our method on our model problem. We measure here the number of outer iterations to achieve an error below $0.05$. We observe an explosion in the number of iterations required for the one-level method to reach the targeted error while our method remains somehow stable. 
\noindent
\begin{minipage}[t]{.55\textwidth}
\begin{figure}[H]
    \centering
     \includegraphics[width=0.9\columnwidth]{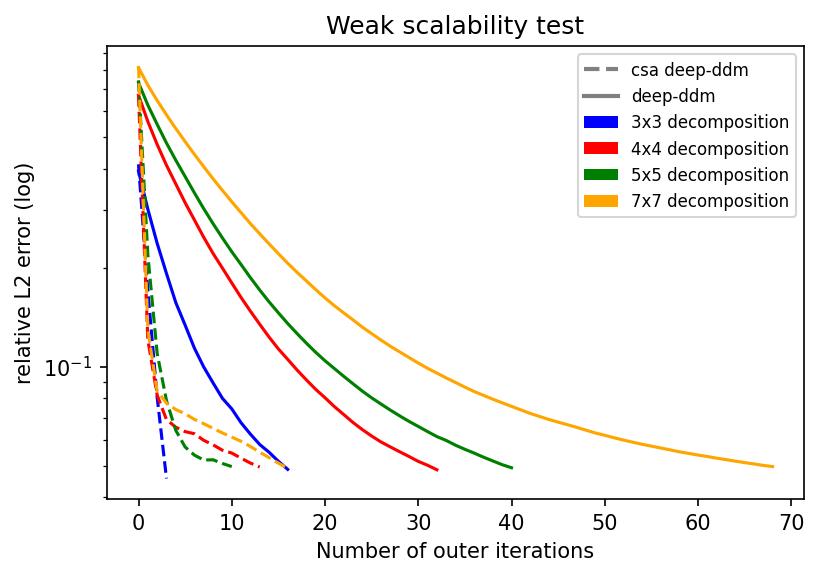}
    \label{fig:weakfig}
    \caption{Convergence history on weak scalability test}
\end{figure}
\end{minipage}
\hfill
\noindent
\begin{minipage}[t]{.45\textwidth}
  \begin{table}[H]
  \centering
  \caption{ Iteration count for a poisson problem. The number of unknowns is proportional to the number of subdomains.}
   \begin{tabular}{@{}c@{}c@{}c}
    \hline
     ~Decomposition~&~Deep-ddm~&~Csa Deep-ddm \\ \hline
     3x3   & 17 & 4 \\
     4x4 &   33 & 14 \\
     5x5&41 & 11 \\
     7x7  & 69 & 17 \\ \hline
     \end{tabular}
     \label{tab:weaktable}
  \end{table}
\end{minipage}
\subsubsection*{Strong scalability}
Using a fix number of sampling points on the original physical problem, we perform a strong scalability test by varying the number of subdomains (and thus of computational units) that are used for the solution of the problem. 
\begin{figure}[H]
    \centering
    \includegraphics[width=0.55\linewidth]{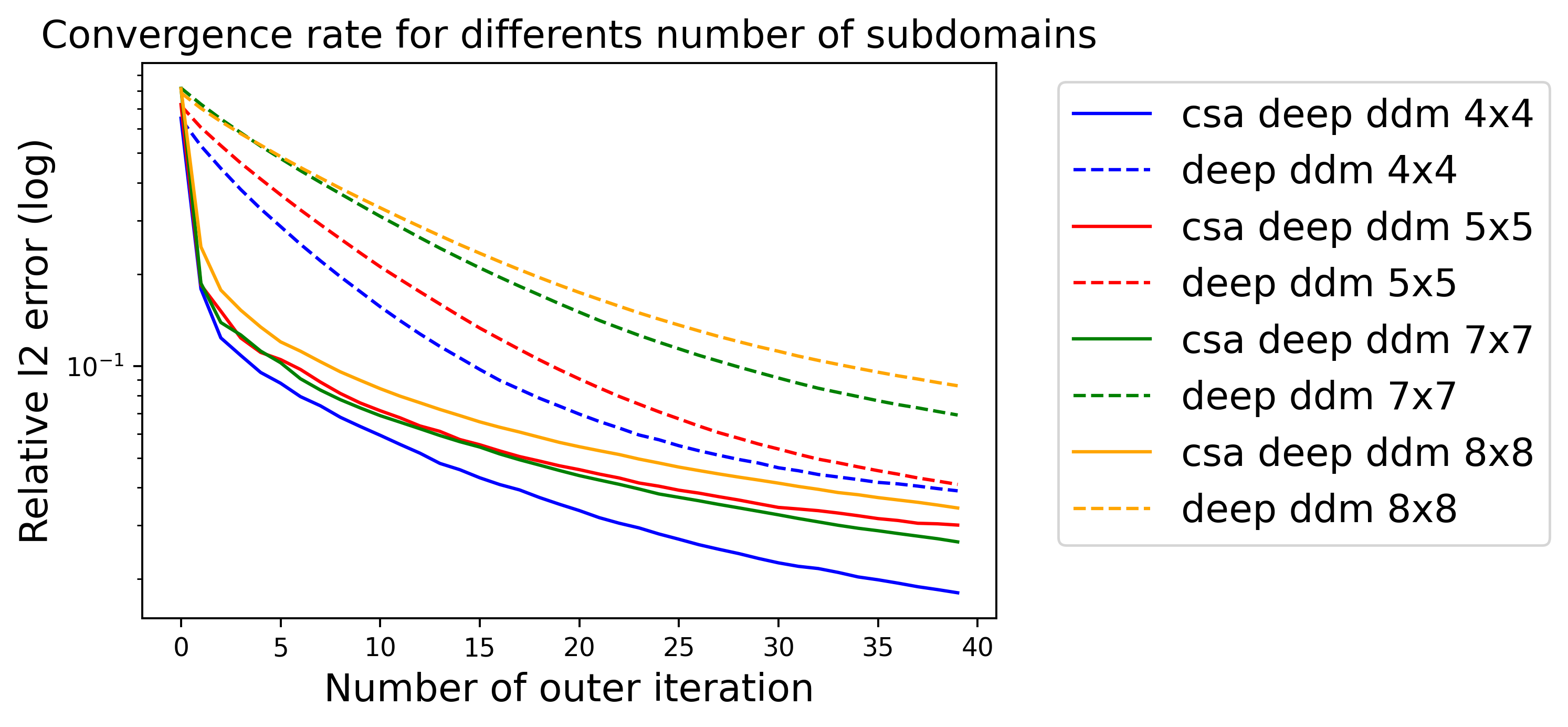}
    \caption{Convergence history without (deep ddm) and with (csa deep-ddm) a coarse network}
    \label{fig:convergencerate}
\end{figure}
We observe a noticeable improvement in convergence, especially on the first iterations. Our method with a 8x8 division converges better than the basic algorithm with a 4x4 division. The dependence on the number of subdomains seems to have decreased, therefore we have improved the strong scalability properties of the algorithm as expected by analogy with DDM. 
\subsection{Information flow}
\begin{wrapfigure}{R}{0.4\linewidth}
    \centering
    \includegraphics[width=\linewidth]{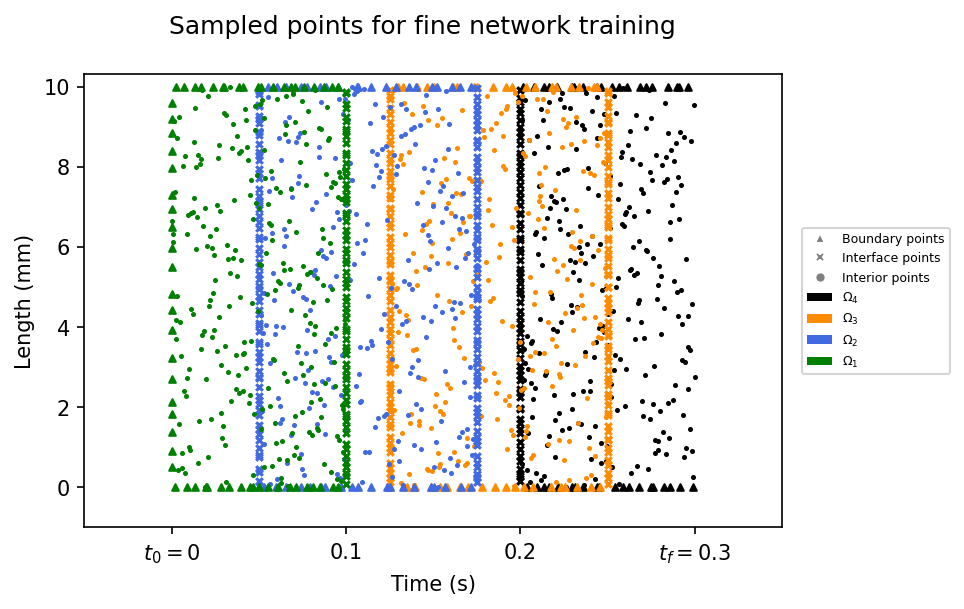}
    \caption{Spatial-temporal decomposition into 4 sub-domains }
    \label{fig:domainechaleur}
\end{wrapfigure}
\begin{figure}[ht]
    \centering
    \subfloat[\centering Error distribution on the presented decomposition (fig \ref{fig:domainechaleur})]{{\includegraphics[width=0.55\linewidth]{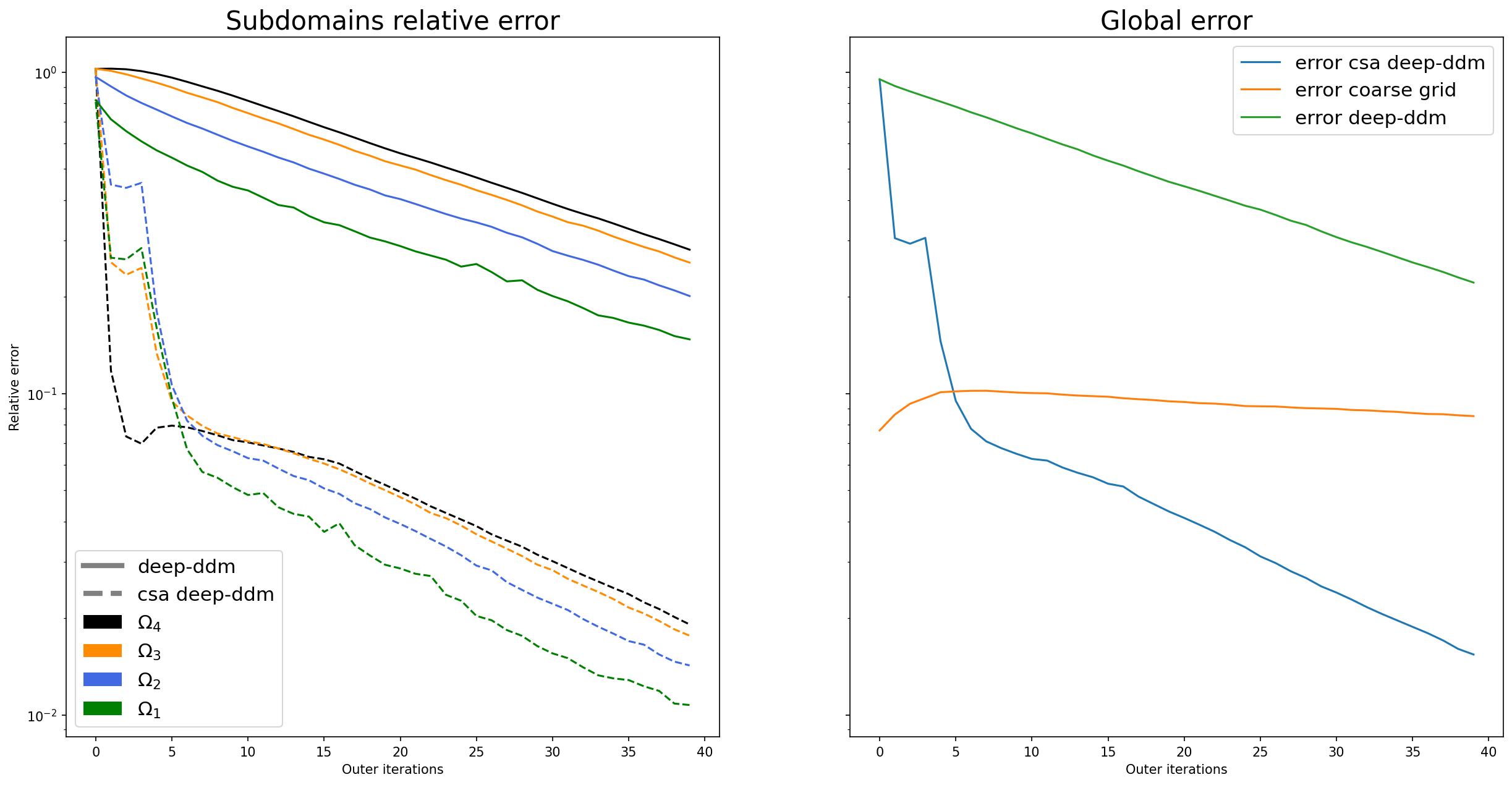} }}%
    \hfill
    \subfloat[\centering Strong scalability test]{{\includegraphics[width=0.42\linewidth]{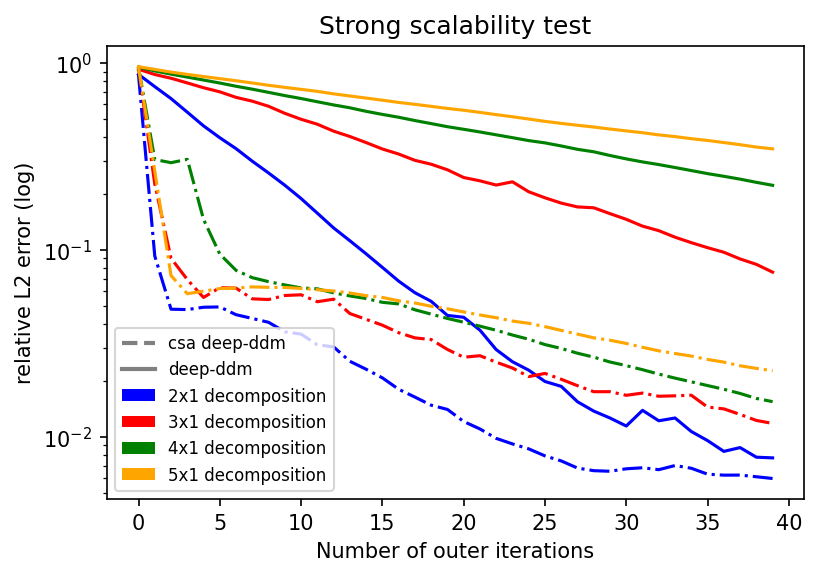} }}%
    \caption{On the left are measurements on the error distribution of the methods in a space-time domain decomposition of the heat equation, on the right a strong scalability test on the same equation.}%
    \label{fig:chaleurerreur}%
\end{figure}

In this section we will work on the 1-d heat equation. This equation is spatio-temporal, so we  use a domain decomposition in time and space. We divide the domain as shown in figure \ref{fig:domainechaleur}. 
We want to illustrate here the information flow in domain decomposition methods, indeed only the domain $\Omega_1$ has a direct access to the initial condition characterizing the solution of our PDE. By observing the errors per sub-domain associated with our method (figure \ref{fig:chaleurerreur}~(a)), we note the strong uniform contribution of the coarse network in the first iterations on all the sub-domains allowing them to converge almost at the same speed. On the other hand, in the deep-ddm method, the time lag between the curves illustrates perfectly the transport time of the information from one domain to the other. We can understand here the effect of the coarse network on our algorithm and its effect on the scalability properties of the method. These effects are confirmed by a strong scalability test with the same kind of decomposition (see figure \ref{fig:chaleurerreur}~(b)).
\section{Conclusion}
We have presented a two-level approach to significantly accelerate the convergence of the original deep-ddm method. The addition of the coarse space and its coarse network has been designed for a better transmission of information and to enhance the scalability properties of deep-ddm methods. We obtained  very promising numerical results on Laplace equation and on the Heat Equation.
Given the way we control the amount of data sampled on the coarse space, the cost of the coarse training is small relative to the total cost of the method. We believe this is an important step towards a broader use of Deep Learning methods for  solving problems involving Physics. However this work is barely a first step.  Many questions remain to be addressed on the hyper-parameters of the method, on the applicability to other differential equation, to mention only a few.

\textbf{Future work and possible improvements :}

As stated in \cite{li_deep_2020}, relying on neural networks raises some fundamental questions: what is the most appropriate sampling for a problem or a network architecture?  What is the most adapted network architecture for a problem? The deep learning community is active on these important questions. 
The work presented is based only on empirical observations and theoretical work will be needed to obtain a deeper understanding. It is worth mentioning that, as stated in \cite{karniadakis_extended_2020}, it will be difficult to achieve relative errors comparable to those obtained with classical methods (e.g. FEM). Indeed, due to the highly non-convex nature of the loss function used, obtaining a global optimum is rather unlikely. 

In the short term, improvements can be made, for example, the use of a standard coarse numerical solver instead of a PINN is a promising avenue and brings us closer to the work of \cite{meng_ppinn_2020} (without however decoupling the problems between them). We also wish to apply the principle of PINNs to the residuals of the physical problem as is done in the classical literature \cite{dolean_introduction_2015} and strengthen the interface conditions using the work already done in \cite{karniadakis_extended_2020} and \cite{jagtap_conservative_2020}. 

\section*{Broader Impact}
The broader impact of deep learning methods for solving PDEs has been detailed in \cite{um_solver---loop_2021}. 
\newpage
\section{Appendix}
\subsection{General information}
All experiments were performed on a 'GeForce GTX 1080 Ti' card in the pytorch framework. 
As a reminder, the neural networks used are fully connected networks of $h$ hidden layers of $w$ neurons interspersed with Tanh activation functions. The parameters of the networks are trained at each outer iteration with an Adam \cite{kingma_adam_2017} optimizer until convergence of the loss functions (or maximum iteration limits) is reached. We will use a learning rate of $10^{-2}$ with a decay between $0.9$ and $0.99$ . 
The training points are all chosen using a Latin hypercube method. The test points used to calculate errors are regularly sampled by row and column, with 4,000,000 in all. We will manipulate problems with known analytical solution and use the relative error $\mathcal{L}^2$.

$$\epsilon = (\frac{\sum_{i=1}^{n}|u_*-u_h|^2}{\sum_{i=1}^{n}|u_*|^2})^{\frac{1}{2}}$$

Below we detail the main hyperparameters used to generate the results presented. 
\subsection{Geometry of the decompositions}
All decompositions presented in this paper are two-dimensional domain decompositions. The decompositions are uniform in $n\times m$ square subdomains. Each subdomain has interior points, interface points and boundary points depending on its position. Here are some examples of decompositions.  

\begin{figure}[!h]
    \centering
    \includegraphics[width=0.7\linewidth]{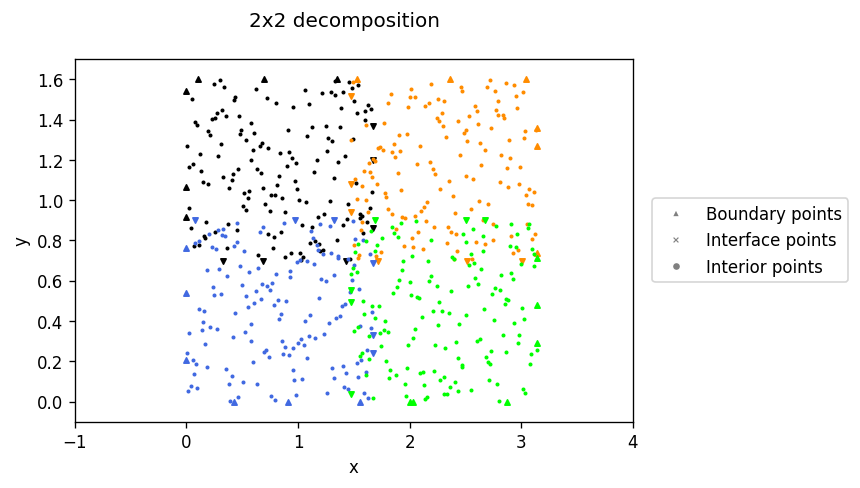}
    \caption{2x2 Decomposition example}
    \label{fig:2x2}
\end{figure}

\begin{figure}[!h]
    \centering
    \includegraphics[width=0.7\linewidth]{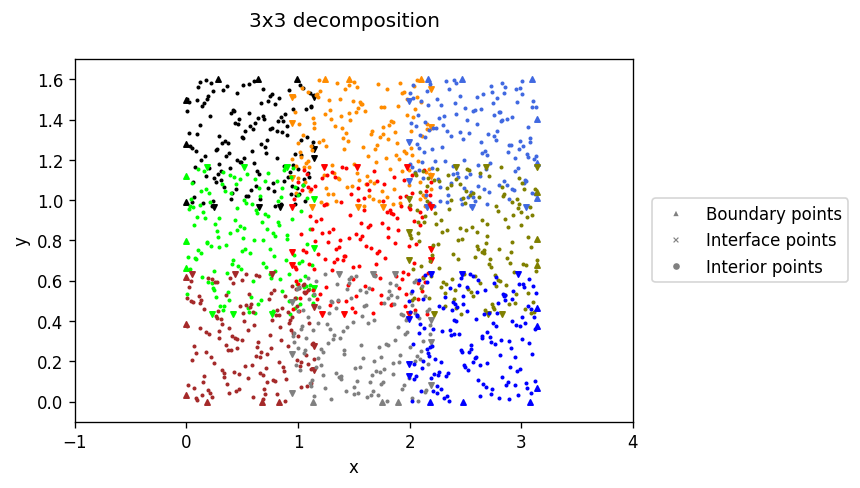}
    \caption{3x3 Decomposition example}
    \label{fig:3x3}
\end{figure}

\begin{figure}[!h]
    \centering
    \includegraphics[width=0.7\linewidth]{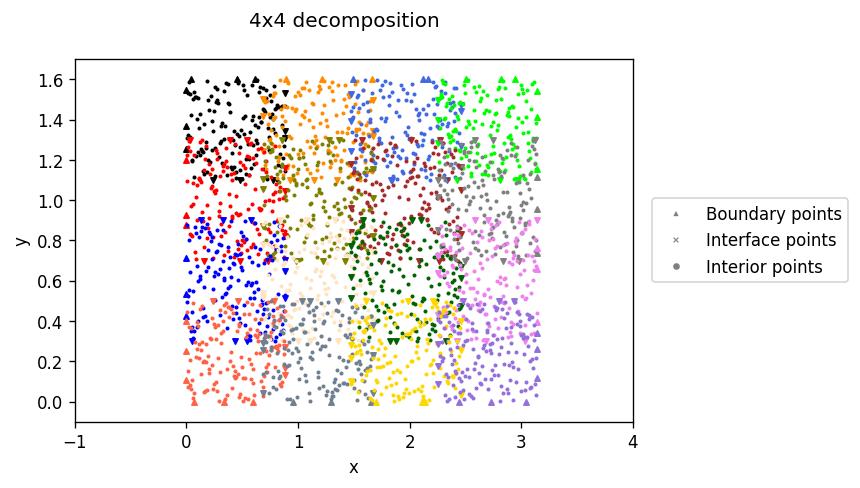}
    \caption{4x4 Decomposition example}
    \label{fig:4x4}
\end{figure}
These examples were generated with the parameters of the experiment 3.
\subsection{Poisson equation}
We consider a Poisson equation with Dirichlet boundary
conditions:
\begin{equation}
\label{general_problem}
    \left\{
    \begin{aligned}
    -\nabla .(\nabla (u))&= f  \text{ in } \Omega  \\
    u(x,y) &= g  \text{ on }  \partial \Omega     
    \end{aligned}
  \right.
\end{equation}

\subsubsection{Experiment 1 (figure 2)}
\label{exp:1}
Here we perform a strong scalability test. The amount of points sampled in the domain remains the same and is distributed equally among all sub-domains. As stated in the conclusion of the paper, there is no method yet to size a PINN according to the equation or the amount of data sampled, so we have chosen to keep the architecture of the models constant. It is possible in the context of a specialised application to adapt the architecture and therefore the approximation strength according to the complexity of the phenomenon in the sub-domain treated, which is a major advantage of the method. The strong scalability graph presented in section 2.4 was performed on a problem presented in the deep-ddm paper. We solve a Poisson equation whose analytical solution is the following: 
$$u_* = sin(2x)e^y$$
with $\Omega =[0,\pi]\times[0,1]$.The following hyperparameters were used : 
\begin{table}[H]
    \centering
    \caption{Hyperparameters figure 2}
    \begin{tabular}{cccccccc}\hline
     $N_f$  & $N_g$ & $N_\Gamma$ & $\delta$ & $tol_m$ & $M_s$\\ \hline
     2500  & 400 & 400 & 0.2 & $1e^{-3}$ & 64\\ \hline
     \end{tabular}
    \label{tab:my_label}
\end{table}
As in the original paper, we used networks with 3 hidden layers of 20 neurons each. 
\subsubsection{Experiment 2 (figure 3)}
For this and the next two experiments, we solve the Poisson equation whose analytical solution is $$u_* = sin(2\pi x)e^y$$  with $\Omega =[0,\pi]\times[0,1.6]$.
In this experiment we test the influence of the coarse network on the convergence of our method. For this purpose we vary the parameter $tol_m^{coarse}$ managing the coarse network training as well as the coarse network contribution parameter $\lambda_c$ as detailed in the paper. We work on a 4x4 domain decomposition with an overlap of 0.2. The sampling of the fine and coarse domain is the same between the generation of two curves, the initializations of the networks are also identical. 
\begin{table}[H]
    \centering
    \caption{Hyperparameters fine networks figure 3}
    \begin{tabular}{cccccccc}\hline
     $N_{f}$  & $N_{g}$ & $N_{\Gamma}$ & $\delta$ & $tol_m$ & $M_s$\\ \hline
     1000  & 80 & 80 & 0.2 & $1e^{-3}$ & 64\\ \hline
     \end{tabular}
    \label{tab:my_label}
\end{table}

\begin{table}[H]
    \centering
    \caption{Hyperparameters coarse network figure 3}
    \begin{tabular}{cccccccc}\hline
     $N_{f,coarse}$  & $N_{g,coarse}$& $tol_m^{coarse}$ & $M_s^{coarse}$ & $\lambda_f$\\ \hline
     200  & 10 & $1e^{-3}$ & 200 & $0.05$\\ \hline
     \end{tabular}
    \label{tab:my_label}
\end{table}
For this test we used fine networks of one hidden layer of twenty neurons and a coarse network of one hidden layer of ten neurons.
\subsubsection{Experiment 3 (figure 4)}
For our weak scalability test, we kept the number of points per subdomain constant (denoted by $N_{fs}$, $N_{gs}$ and $N_{\Gamma s}$) and then varied the number of subdomains used. For the same reasons as the previous test, we keep the network architecture fixed for both the fine and coarse networks. In order to be able to compare the two methods, the same fine samplings and the same fine network initializations were used for a given decomposition. The same coarse network initialization was used for all decompositions. 

\begin{table}[H]
    \centering
    \caption{Hyperparameters fine networks figure 4}
    \begin{tabular}{cccccccc}\hline
     $N_{fs}$  & $N_{gs}$ & $N_{\Gamma s}$ & $\delta$ & $tol_m$ & $M_s$\\ \hline
     144  & 3 & 3 & 0.2 & $1e^{-3}$ & 64\\ \hline
     \end{tabular}
    \label{tab:my_label}
\end{table}

\begin{table}[H]
    \centering
    \caption{Hyperparameters coarse network figure 4}
    \begin{tabular}{cccccccc}\hline
     $N_{f,coarse}$  & $N_{g,coarse}$ & $tol_m^{coarse}$ & $M_s^{coarse}$ & $\lambda_c$ & $\lambda_f$ \\ \hline
     144  & 4 & $1e^{-3}$ & 144& $0.9*0.8^{outerit}$& $0.05$\\ \hline
     \end{tabular}
    \label{tab:my_label}
\end{table}
For this test we used fine networks of one hidden layer of twenty neurons and a coarse network of one hidden layer of ten neurons. 

\textcolor{red}{\textbf{Typo :}}
There is a typo in the graph in Figure 4. The solid lines represent the convergence history of the deep-ddm method unlike what is written in the legend.  

\subsubsection{Experiment 4 (figure 5)}
We again perform a strong scalability test, as in the previous experiment the fine samplings and fine network intializations are the same for both methods (deep-ddm and csa deep-ddm). The coarse sampling and coarse network intialization also remain constant between all decompositions. 
\begin{table}[H]
    \centering
    \caption{Hyperparameters fine networks figure 5}
    \begin{tabular}{cccccccc}\hline
     $N_{f}$  & $N_{g}$ & $N_{\Gamma}$ & $\delta$ & $tol_m$ & $M_s$\\ \hline
     5000  & 400 & 400 & 0.2 & $1e^{-3}$ & $N_{f s}/2$\\ \hline
     \end{tabular}
    \label{tab:my_label}
\end{table}

\begin{table}[H]
    \centering
    \caption{Hyperparameters coarse network figure 5}
    \begin{tabular}{cccccccc}\hline
     $N_{f,coarse}$  & $N_{g,coarse}$& $tol_m^{coarse}$ & $M_s^{coarse}$ & $\lambda_c$ & $\lambda_f$\\ \hline
     400  & 5 & $1e^{-3}$ & 400 & $0.9*0.8^{outerit}$& $0.05$\\ \hline
     \end{tabular}
    \label{tab:my_label}
\end{table}
For this test we used fine networks of one hidden layer of twenty neurons and a coarse network of one hidden layer of ten neurons.

\subsection{Heat equation}
The following experiments were performed on the heat equation with Dirichlet boundary conditions. 
\begin{equation}
\label{general_problem}
    \left\{
    \begin{aligned}
    \frac{\partial T}{\partial t} &=  \alpha \frac{\partial^2 T}{\partial x^2} \text{ in } \Omega  \\
    T(x_{edge},t) &= T_{edge}(x_{edge},t) \text{ on } \Omega_{edge}\\
    T(x,0) &= T_{init}(x,0) \text{ on } \Omega_{init}\\
    \end{aligned}
  \right.
\end{equation}
in which $\alpha$ is a positive coefficient called the thermal diffusivity of the medium. We define for our experiments $\Omega = [0,10]\times[0,0.3], \quad \Omega_{edge} = \{0,1\}\times[0,0.3], \quad \Omega_{init}=[0,10]\times\{0\}  \text{ and } \alpha=4$. 
The domain is heated according to a Gaussian at $t=0$, the edges are at $0^{\circ}C$ for all $t$. 
To obtain numerical results for comparison, we use a finite difference method.We will generate 24000 comparison points with 100 spatially discretised points and 240 time steps. 
\subsubsection{Experiment 5 (figure 7(a))}
As detailed in the paper, the objective of the experiment is to illustrate the time taken to transmit information between sub-domains. We measure here the error per subdomain of both methods and analyse the shift between the convergence curves. As for the previous experiments, the initialization of the networks and the sampling are the same for both methods. We break down our domain into three sub-domains as presented in the article. 
\begin{table}[H]
    \centering
    \caption{Hyperparameters fine networks figure 7(a)}
    \begin{tabular}{cccccccc}\hline
     $N_{f}$  & $N_{g}$ & $N_{\Gamma}$ & $\delta$ & $tol_m$ & $M_s$\\ \hline
     1000  & 200 & 200 & 0.05 & $1e^{-3}$ & $N_{f s}/2$\\ \hline
     \end{tabular}
    \label{tab:my_label}
\end{table}

\begin{table}[H]
    \centering
    \caption{Hyperparameters coarse network figure 7(a)}
    \begin{tabular}{cccccccc}\hline
     $N_{f,coarse}$  & $N_{g,coarse}$& $tol_m^{coarse}$ & $M_s^{coarse}$ & $\lambda_c$ & $\lambda_f$\\ \hline
     60  & 6 & $1e^{-3}$ & 60 & $0.9*0.8^{outerit}$& $0.05$\\ \hline
     \end{tabular}
    \label{tab:my_label}
\end{table}
For this test we used fine networks of one hidden layer of twenty neurons and a coarse network of one hidden layer of ten neurons.
\subsubsection{Experiment 6 (figure 7(b))}
To generate the graph, we carry out a strong scalability test with the same sampling and initialisation methods as the previous tests. It is important to note that the decomposition will only be done on one axis (time) as shown in the example below. 

\begin{table}[H]
    \centering
    \caption{Hyperparameters fine networks figure 7(b)}
    \begin{tabular}{cccccccc}\hline
     $N_{f}$  & $N_{g}$ & $N_{\Gamma}$ & $\delta$ & $tol_m$ & $M_s$\\ \hline
     1000  & 200 & 200 & 0.05 & $1e^{-3}$ & $N_{f s}/2$\\ \hline
     \end{tabular}
    \label{tab:my_label}
\end{table}

\begin{table}[H]
    \centering
    \caption{Hyperparameters coarse network figure 7(b)}
    \begin{tabular}{cccccccc}\hline
     $N_{f,coarse}$  & $N_{g,coarse}$& $tol_m^{coarse}$ & $M_s^{coarse}$ & $\lambda_c$ & $\lambda_f$\\ \hline
     60  & 6 & $1e^{-3}$ & 60 & $0.9*0.8^{outerit}$& $0.05$\\ \hline
     \end{tabular}
    \label{tab:my_label}
\end{table}

\begin{figure}[H]
    \centering
    \includegraphics[width=\linewidth]{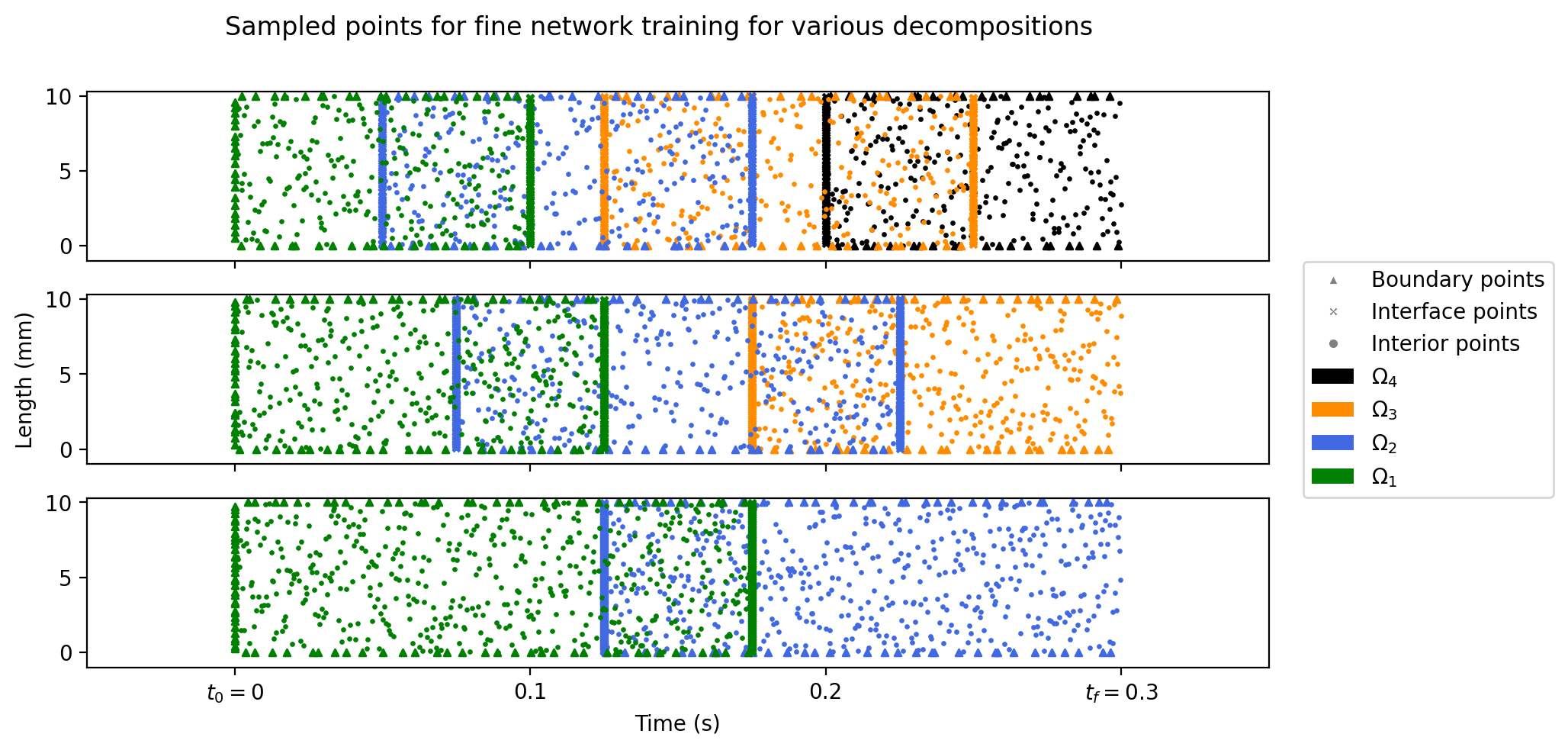}
    \caption{Different domain decomposition for strong scalability test on the heat equation. }
    \label{fig:decompo}
\end{figure}
\subsection{Glossary}
\begin{itemize}
    \item $N_f$ : Number of interior points sampled on the domain 
    \item $N_g$ : Number of boundary points sampled on the domain 
    \item $N_\Gamma$ : Number of points at the interfaces sampled on the domain 
    \item $N_{fs}$ : Number of interior points sampled on the domain on the sub-domain $s$
    \item $N_{gs}$ : Number of boundary points sampled on the domain on the sub-domain $s$
    \item $N_{\Gamma s}$ : Number of points at the interfaces sampled on the domain on the sub-domain $s$
    \item $\delta$ : overlap between two sub-domains
    \item $tol_m$ : stop criteria for fine network training
    \item $M_s$ : fine batch construction parameter (detail in \cite{li_deep_2020})
    \item $\lambda_c$ weight of the influence of the coarse network in the composition of the values at the interfaces (details in the paper section 3)
    \item $\lambda_f$ : Weight of the influence of fine networks on the coarse network through the loss $\mathcal{M}_{fine}$ (details in the paper section 3)
    \item $N_{f,coarse}$ : Number of boundary points sampled on the coarse domain 
    \item $N_{g,coarse}$ : Number of boundary points sampled on the coarse domain 
    \item $M_s^{coarse}$ : coarse batch construction parameter 
    \item deep-ddm : the original method \cite{li_deep_2020}
    \item csa deep-ddm : our coarse space accelerated methods 
\end{itemize}
\bibliography{main.bib}
\bibliographystyle{plain}
\end{document}